\documentclass[lettersize,journal]{IEEEtran}
\usepackage{amsmath,amsfonts}
\usepackage{algorithmic}
\usepackage{algorithm}
\usepackage{array}

\usepackage[caption=false,font=footnotesize,labelfont=sf,textfont=sf]{subfig}
\usepackage{textcomp}
\usepackage{stfloats}
\usepackage{url}
\usepackage{verbatim}
\usepackage{graphicx}
\usepackage{cite}

\begin{document}

\title{MEStereo-Du2CNN: A Novel Dual Channel CNN for Learning Robust Depth Estimates from Multi-exposure Stereo Images for HDR 3D Applications}

\author{Rohit Choudhary $^{1}$,~Mansi~Sharma $^{1}$, ~Uma T V $^{2}$ and ~Rithvik Anil $^{2}$
\thanks{$^1$Department of Electrical Engineering,
Indian Institute of Technology Madras, Tamil Nadu 600036, India
(e-mail: ee20s002@smail.iitm.ac.in, mansisharmaiitd@gmail.com, mansisharma@ee.iitm.ac.in)}

\thanks{$^2$Department of Mechanical Engineering, Indian Institute of Technology Madras, Tamil Nadu 600036, India (e-mail: me17b170@smail.iitm.ac.in, rithvik.anil@gmail.com)}
}

\maketitle

\begin{abstract}
Display technologies have evolved over the years. It is critical to develop
practical HDR capturing, processing, and display solutions to bring 3D technologies to the next level. Depth estimation of multi-exposure stereo image sequences is an essential task in the development of cost-effective 3D HDR video content. In this paper, we develop a novel deep architecture for multi-exposure stereo depth estimation. The proposed architecture has two novel components.
First, the stereo matching technique used in traditional stereo depth estimation is revamped. For the stereo depth estimation component of our architecture, a mono-to-stereo transfer learning approach is deployed. The proposed formulation circumvents the cost volume construction requirement, which is replaced by a ResNet based dual-encoder single-decoder CNN with different weights for feature fusion. EfficientNet based blocks are used to learn the disparity. Secondly, we combine disparity maps obtained from the stereo images at different exposure levels using a robust disparity feature fusion approach. The disparity maps obtained at different exposures are merged using weight maps calculated for different quality measures. The final predicted disparity map obtained is more robust and retains best features that preserve the depth discontinuities. The proposed CNN offers flexibility to train using standard dynamic range stereo data or with multi-exposure low dynamic range stereo sequences. In terms of performance, the proposed model surpasses state-of-the-art monocular and stereo depth estimation methods, both quantitatively and qualitatively, on challenging Scene flow and differently exposed Middlebury stereo datasets. The architecture performs exceedingly well on complex natural scenes, demonstrating its usefulness for diverse 3D HDR applications. 
\end{abstract}


\section{Introduction}
\label{sec:introduction}

With the development of advanced visual technologies such as Augmented Reality, Virtual Reality, Autostereoscopic Glasses-free 3D Displays, etc., there is an increasing demand for high-quality 3D video content. High Dynamic Range (HDR) 3D video technology has gained popularity over the last few years. Current cameras and displays can span over a standard dynamic range (contrast) of 300:1 to 1,000:1. However, the human visual system can adapt to a much larger dynamic range of 50,000:1 or more \cite{HDRImaging2015}. HDR videos can produce a dynamic range very near to the Human Visual System (HVS). Consequently, HDR video provides a more realistic experience depicting reflection, refraction, specularities, and global illumination effects.

HDR image acquisition is either done using expensive HDR cameras or through HDR image reconstruction from the visual content captured by low-dynamic range cameras \cite{Wang2021DLforHDRI}. Due to the ease of implementation and reduced cost compared to HDR cameras, HDR image reconstruction is preferred by companies that produce consumer-grade products. There are two methods commonly used for HDR image reconstruction using standard dynamic range (SDR) images. The first method involves combining several SDR images of the same scene taken at various exposure times to create HDR content \cite{Yan2020DHDRINonLocal}, \cite{Kalantari2017DHDRIDynScenes}. The second method involves creating HDR content from a single-exposure SDR image \cite{Eilertsen2017HDRIReSingleEx}, \cite{Liu2020SImageHDR}, \cite{Ning2018LearningInvTM}. Industries and research communities have been showing increasing attention to the convergence of 3D and HDR technologies for immersive, high-quality viewing experiences on a variety of display devices. Recently, there has been a solid push to generate cost-effective 3D HDR content. 3D HDR application requires not only the HDR image, but also the scene depth. While many algorithms are available for HDR image reconstruction,  robust depth estimation is still a challenging task for developing 3D HDR video content from multi-exposure stereo datasets acquired with dual camera setups. Keeping the HDR image reconstruction in mind through differently exposed SDR images, we focus on developing a novel and efficient multi-exposure stereo depth estimation framework for robust 3D HDR content generation.

Most of the existing state-of-the-art monocular and stereo-based depth prediction methods are designed or tested on SDR images or videos \cite{DenseDepth2018,DepthHints2019,FCRN2016,SerialUNet2020,MSDN2014,2019RevisitingSIDE,MiDaS2020,Adabins2021,CADepth2021,DeepPruner2019,HSMNet2019,PSMNet2018,STTR2020}. Due to the limited dynamic range of SDR camera sensor, the acquired image of a real-world scene contains under- and over-exposed regions. Such regions do not have adequate information about the texture and thus lack details. Existing stereo matching algorithms output erroneous depth values in such low-textured areas \cite{DeepPruner2019,HSMNet2019,PSMNet2018,STTR2020}.
Some approaches combine high dynamic range (HDR) images with stereo matching techniques to obtain the disparity maps \cite{Akhavan2013AFF,Akhavan2015BackwardCH}. 
The robust depth estimation remains an ill-posed problem in many scenarios, despite attempts to adapt existing stereo matching techniques for HDR and multi-exposed scenes. Inferring consistent depth from multi-exposure stereo views of natural scenes is even more difficult on account of change in visibility due to viewpoint variation, change of illumination, natural lighting, non-Lambertian reflections or partially transparent surfaces, scale variations, the influence of low-textured regions, high details and discontinuities in natural structures. The stereo algorithms are more susceptible to subpixel calibration errors and dependent on the scene complexity. The large mismatches may produce erroneous results in complex natural scenes \cite{DenseDepth2018,DepthHints2019,FCRN2016,SerialUNet2020,MSDN2014,2019RevisitingSIDE,MiDaS2020}.

Typically, the stereo depth estimation pipeline involves four main steps, (1) extraction of the feature, (2) feature matching, (3) disparity estimation, and (4) refining of the acquired disparity. Initially, the features are generated by the convolutional neural network using stereo images. The feature matching is performed by calculating a similarity score at a number of disparity levels. The cost volume is calculated in computing the similarity score based on different metric measures. This generates a 3D or 4D cost volume tensor, which is then used to anticipate and enhance the depth map via a series of convolutions. However, the problem in monocular depth estimation is formulated as a direct regression to the depth space from the image space \cite{DenseDepth2018}.

In this paper, we propose a novel architecture, dubbed as MEStereo-Du2CNN, that addresses challenging depth estimation problems using multi-exposed stereo images for 3D HDR applications. Our proposed model disseminates following novel ideas: 
\vskip 0.05in

\begin{itemize}

\item{We have introduced a mono-to-stereo transfer learning module in MEStereo-DU2CNN to help facilitate the process of stereo depth estimation using progress gained in the monocular depth estimation domain. This is accomplished by feeding the network monocular depth clues at various exposure levels. The module allows the encoded version of each view to provide descriptive features pertaining specifically to depth estimation. }

\vskip 0.05in

\item{The Dual-Channel CNN component in our proposed architecture circumvents the cost volume construction requirement for the stereo matching task. It replaces the explicit data structure, i.e., cost volume, with a combination of ``allowance for different weights in the dual encoders'' and a ``novel element-wise multiplication-based fusion strategy for features from the dual encoders before sending them to the decoder.'' This component better handles dynamic range locally and globally for predicting disparity.}

\vskip 0.05in

\item{Our proposed architecture employs a novel disparity map fusion approach to obtain a refined disparity map by fusing the disparity estimates corresponding to the multi-exposure stereo pairs. Weights obtained from two quality measures: contrast and well-exposedness, are used in the fusion process. The disparity maps, acquired as output to the dual-channel architecture, provide weight for the contrast measure, while the multi-exposure input images provide weight for the well-exposedness measure. These quality measures help in achieving a refined disparity map prediction by retaining the best features that preserve the depth discontinuities.}

\vskip 0.05in

\item{The flexibility of our proposed architecture in terms of its applicability is itself broad and novel. For the process of HDR image reconstruction using multi-exposure SDR images, both exposure fusion and HDR synthesis can be realized in the encoder depending on the availability of the HDR displayer. Considering the HDR image reconstruction through differently exposed SDR images, we have proposed MEStereo-Du2CNN architecture to estimate the scene depth using multi-exposure SDR input. The framework is flexible as the estimated depth maps find their application on both LDR/SDR displays and HDR displays. Thus, the same framework can work for displaying 3D LDR/SDR and also 3D HDR content depending on the display type/application scenario. Additionally, by considering multi-exposed SDR images as inputs for scene depth estimation, our method bypasses the complex process of depth generation from floating-point values in HDR data.}
\end{itemize}

Our architecture replaces two components of traditional stereo depth estimation approaches, i.e., the cost volume construction and encoders with shared weight, with a novel ResNet based dual-encoder single-decoder framework using different weights. Also, ConvNet based blocks in the encoders are replaced by EfficientNet based blocks. The features in the network are fused element-wise at multiple resolutions and then passed to the decoder. The operations of feature fusion and back-propagation are accountable for capturing the stereo information through the encoder weights. The features of stereo images are shifted at each disparity level in traditional approaches to construct cost volume, requiring a maximum disparity value for feature shifting. In our architecture, the network is allowed to learn the maximum disparity value by itself, and this produces more robust results. 

A shorter conference version to lay the foundation of this work is published at IEEE VCIP 2021 \cite{SDE-DualENet}. In this journal paper, we are extending the algorithm for the challenging multi-exposure stereo depth estimation problem. There are two major new components: 1) adaption of a mono-to-stereo transfer learning approach for multi-exposure stereo depth estimation, and 2) a robust disparity fusion component based on extraction of weight maps obtained from different quality measures. It includes an extensive analysis of the performance of the proposed CNN on multi-exposure stereo data sequences supported by detailed results. The proposed extension aims at robust depth estimation for 3D HDR applications. The rest of this article is divided into four major sections. Section \ref{sec:RelatedWork} discusses various image-based depth estimation algorithms. The proposed CNN architecture is described in detail in Section \ref{sec:ProposedCNNArchitecture}. In Section \ref{sec:ExperimentsAndResults}, we elaborate our experiments describing the implementation, results, and detailed analysis. Finally, Section \ref{sec:Conclusion}
presents the conclusion of proposed scheme with comprehensive findings and
implications of future work.

%Figure 1
\begin{figure*}[h!]
\centering
{\includegraphics[height=3.7in, width=6.8in]{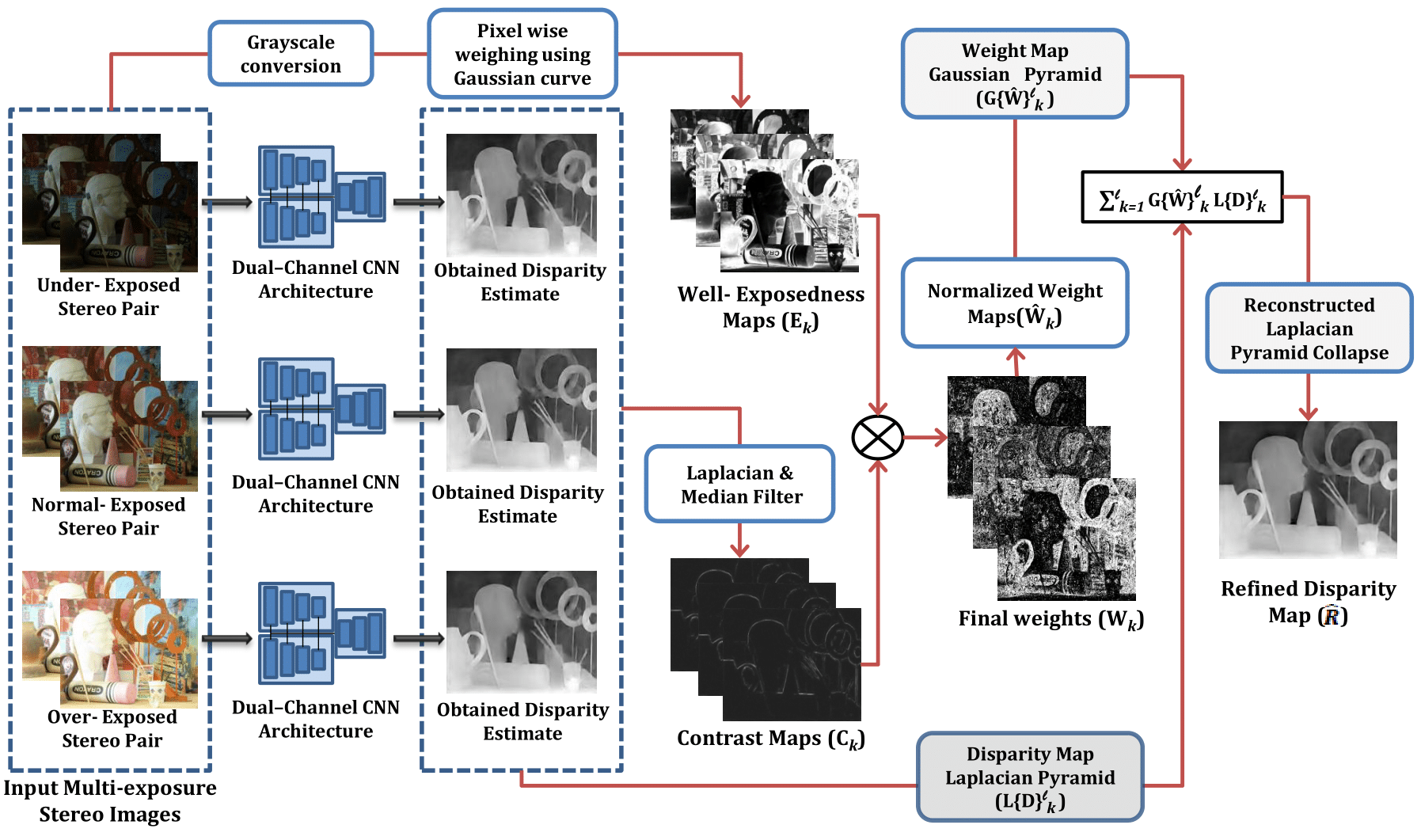}}\hfill
\caption{Overview of proposed ME2Stereo-Du2CNN architecture: The input consists of three stereo pair of the same scene captured at different camera exposure level and the output is a refined disparity map $(\hat{R})$. The input is passed through novel Dual CNN architecture to get the disparity estimates at three exposure levels. The estimates are fused using two quality measures: well-exposedness and contrast, which in-turn are obtained respectively from the input left-view image and the disparity estimate of the corresponding stereo pair. The predicted refined disparity map $(\hat{R})$ is a better estimate compared to three intermediate disparity estimates.}

\label{MEStereoDu2CNNArc}
\end{figure*}

\section{Related Work}
\label{sec:RelatedWork}
In this section, we give a brief review of studies for image-based depth estimation.

\subsection{Monocular Depth Estimation}
Several CNN methods have considered monocular depth estimation, where the problem is posed as a regression of depth map from a single RGB image \cite{MSDN2014, FCRN2016, ContinousCRFCNN2017, SIuAGNs2018, SAGCNF2018}.

Eigen et al. \cite{MSDN2014} combined both local and global cues by using two stack Deep networks for the task of monocular depth estimation. The first stack makes coarse prediction globally based on the whole image, while the second refines it locally. Using different encoder and decoder architectures, Alhashim et al.\cite{DenseDepth2018} showed that increasing complexity of convolutional blocks doesn't necessarily improve the performance of architecture for the task of depth estimation. Thus, it is possible to achieve high resolution depth maps using a simple encoder-decoder architecture with proper augmentation policy and training strategies. They proposed a convolutional neural network to get high resolution depth map of a scene using transfer learning.

When estimating the depth of a scene, the loss of spatial resolution results in distorted object boundaries and absence of minute details from the depth maps. Hu et al. \cite{2019RevisitingSIDE} proposed two enhancement techniques to the existing depth estimation approaches for obtaining high spatial resolution depth maps. One, applying a fusion strategy for obtaining features at different scales. The other one is minimizing inference errors during training using an improved loss function. Ranftl el al. \cite{MiDaS2020} showed that mixing data from complementary sources for the task of model training, considerably improves the monocular depth estimation of diverse real scenes. They targeted important training objectives invariant to depth range and scale variations. They advocated the use of principled multi-objective learning and the importance of pre-training encoders for auxiliary tasks.  

Watson et al. \cite{DepthHints2019} examined the issue of re-projection in depth prediction from stereo-based self-supervision. They reduced this effect by introducing complementary depth suggestions, termed as Depth Hints. Liana et al. \cite{FCRN2016} proposed a powerful, single-scale CNN architecture accompanying residual learning. Cantrell et al. \cite{SerialUNet2020} aimed at integrating the advantages of transfer learning and semantic segmentation for better depth estimation results. Bhat et al. \cite{Adabins2021} introduced a new transformer-based architectural block, dubbed as AdaBins for the task of depth estimation from a single image. The block separates the depth ranges into bins each with an adaptively calculated center value. A linear combination of the center of bins gives the final estimated value of the depth.

%FIGURE 2
\begin{figure*}[h!]
\centering
{\includegraphics[height =2.3in, width=6.7in]{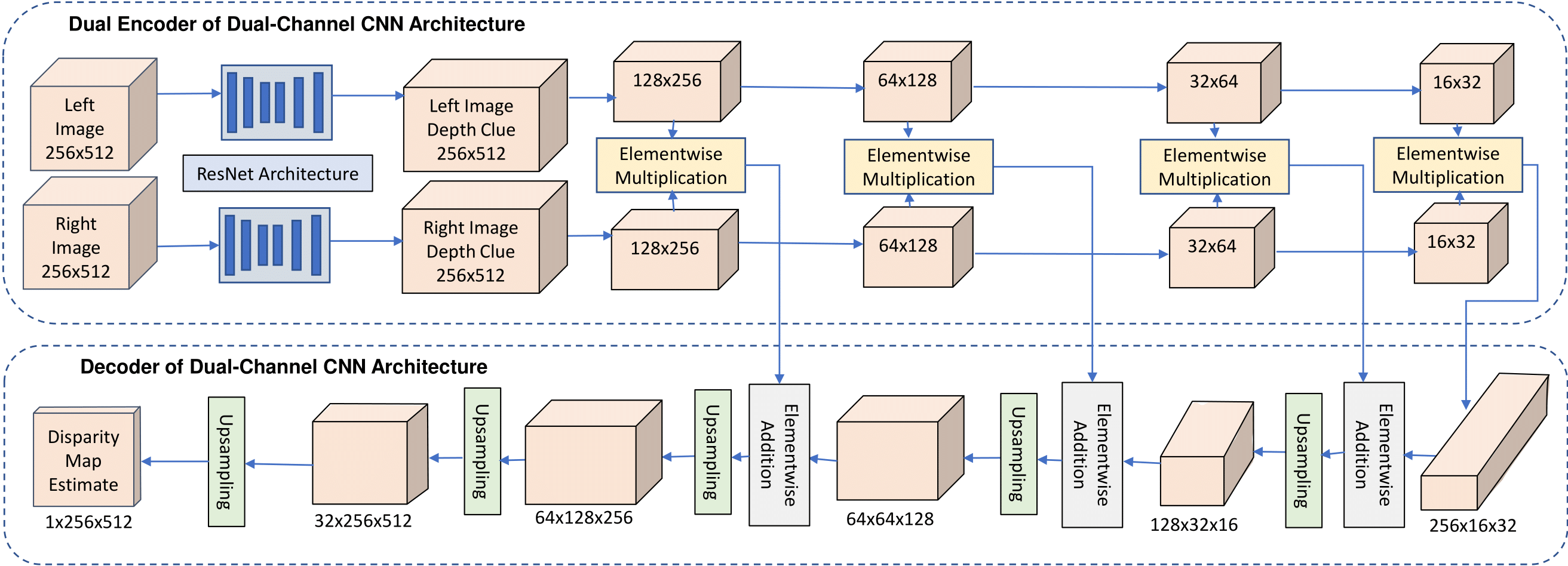}}\hfill
%\caption{MEStereo-Du2CNN Architecture – This figure describes proposed Dual Channel CNN used for stereo depth estimation.}
\caption{A schematic representation of novel Dual-Channel CNN Architecture: It consists of dual parallel encoder followed by a single decoder. The network takes a stereo pair as input and outputs the disparity map estimate. The left and the right views are fed into ResNet architecture to obtain the respective monocular depth clues, which are then passed to the encoders. Our network uses a simple element-wise multiplication of the features at multiple resolutions, which in-turn is fed into decoders at the corresponding resolutions. During back-propagation, the weights of dual encoders are shifted in order to capture the stereo information and this is used to obtain the final disparity.}
\label{Du2CNNArc}
\end{figure*}

Yan et. al \cite{CADepth2021} proposed a channel-wise attention-based depth estimation network with two effective modules to efficiently handle the overall structure and local details. The structural perception module aggregates the discriminative features by capturing the long-range dependencies to obtain the context of scene structure and rich feature representation. The detail emphasis module employs the channel attention mechanism to highlight objects’ boundaries information and efficiently fuse various level features.

\subsection{Stereo Depth Estimation}

Depth estimation from stereo images generally includes three phases \cite{TaxEvalStereoCorres2001}: calculation of a pixel-wise feature representation, the cost volume construction, and a final post-processing. The stereo matching problem is traditionally tackled using dynamic programming approaches, where matching is computed using pixel intensities and costs are aggregated horizontally in 1D \cite{IntraInterDP1985} or multi-directionally in 2D \cite{SMsemiglobal2008}. Networks such as \cite{KendallDeepStereoRegression2017} learn to concatenate features with varied disparities to form a 4D feature volume,  then compute the matching cost using 3D convolutions. Modern approaches use CNNs to extract robust features and execute matching in order to deal with increasingly complicated real-world scenarios, such as texture-less areas or reflecting surfaces. Methods like \cite{SFFlying3D},\cite{LiangFeatureConstancy2018} used learning-based feature extractors to calculate similarities between each pixel’s feature descriptors.

Yang et al. \cite{HSMNet2019} addressed the speed and memory constraints while computing the depths of high resolution stereo images. They used a hierarchical stereo matching architecture that initially down-sample the high resolution images, while extracting the multi-scale features followed by utilizing  potential correspondences to build up the cost volumes pyramid that increases in resolution. 

To overcome the difficulty of finding the exact corresponding points in inherently ill-posed regions, Chang et al. \cite{PSMNet2018} proposed a pyramid stereo matching network consisting of two main modules. The Spatial Pyramid Pooling module incorporates global context information into image features, and 3D CNN module extends the regional support of context information in cost volume. Li et al. \cite{STTR2020} used the position information and attention with respect to the stereo images to replace the cost volume construction with dense pixel matching.

Most stereo matching algorithms usually generate a cost volume over the full disparity space, which increases the computation burden as well as the memory consumption. Duggal et al. \cite{DeepPruner2019} considered speeding up the stereo depth estimation real-time inference by pruning the part of cost volume for each pixel without fully evaluating the related matching score, using a Differential PatchMatch module.

\subsection{HDR Depth Estimation}
Akhavan et al. \cite{Akhavan2013AFF} proposed a theoretical framework with three possible approaches, determining the depth map using multi-exposed stereo images (under-, normal- and over-exposed) with respect to a scene. The first approach involves constructing HDR images for both (left and right) views, followed by computation of disparity map between two HDR images. The second approach uses a tone mapper to convert the HDR stereo pair into a perceptually low dynamic range stereo pair. After that, a suitable stereo matching algorithm is applied to the tone-mapped stereo pair. In the third approach, disparity maps are calculated for stereo pairs corresponding to different exposure levels. They suggested a fuzzy measure and integral combination method with respect to the third approach, to achieve an accurate disparity map from different exposures. Likewise, Akhavan and Kaufmann \cite{Akhavan2015BackwardCH} presented a backward compatible stereo matching method for HDR scenes. The disparity maps from different tone mapped stereo images are effectively fused through a graph cut based framework.

Scenes captured under low light conditions exhibit low image quality and imprecise depth acquisition. Im et al. \cite{Im:AutoExpBracketing} proposed 
a narrow baseline multi-view stereo matching method that delivers a robust depth estimation for a short burst shot with altering intensity. The authors determined to use the unavoidable motion occurring during shutter capture in burst photography as an important clue to estimate the depth from a short burst shot with varied intensity. They presented a geometric transformation between the optical flow and depth of the burst images, exploiting the geometric information of the scene, such as camera poses and sparse 3D points. This is incorporated within the residual flow network. In another approach, Yung et al. \cite{HueiYung2016} modified the existing state-of-the-art stereo matching algorithms and make them compatible to HDR scenes with image storage slice down to 16 bits per channel.

Chari et al. \cite{Chari2020OptimalHA} generalized the noise optimal framework by Hasinoff et al. \cite{Hasinoff2010} for determining the best exposure and optimal ISO sequence for HDR recovery and depth reconstruction from a dual-camera setup. Multi-exposure LDR image sequence is used as an input to estimate inverse camera response functions (ICRFs), scene disparity maps, and HDR images. They employed the Mozerov et al. \cite{Mozerov2015SM2StepEnergyMin} disparity estimation algorithm to demonstrate disparity output using their framework, which is adaptable to different disparity estimation algorithms. 

%FIGURE 3
\begin{figure}[t!]
\centering
{\includegraphics[height = 2.05 in, width=3.3in]{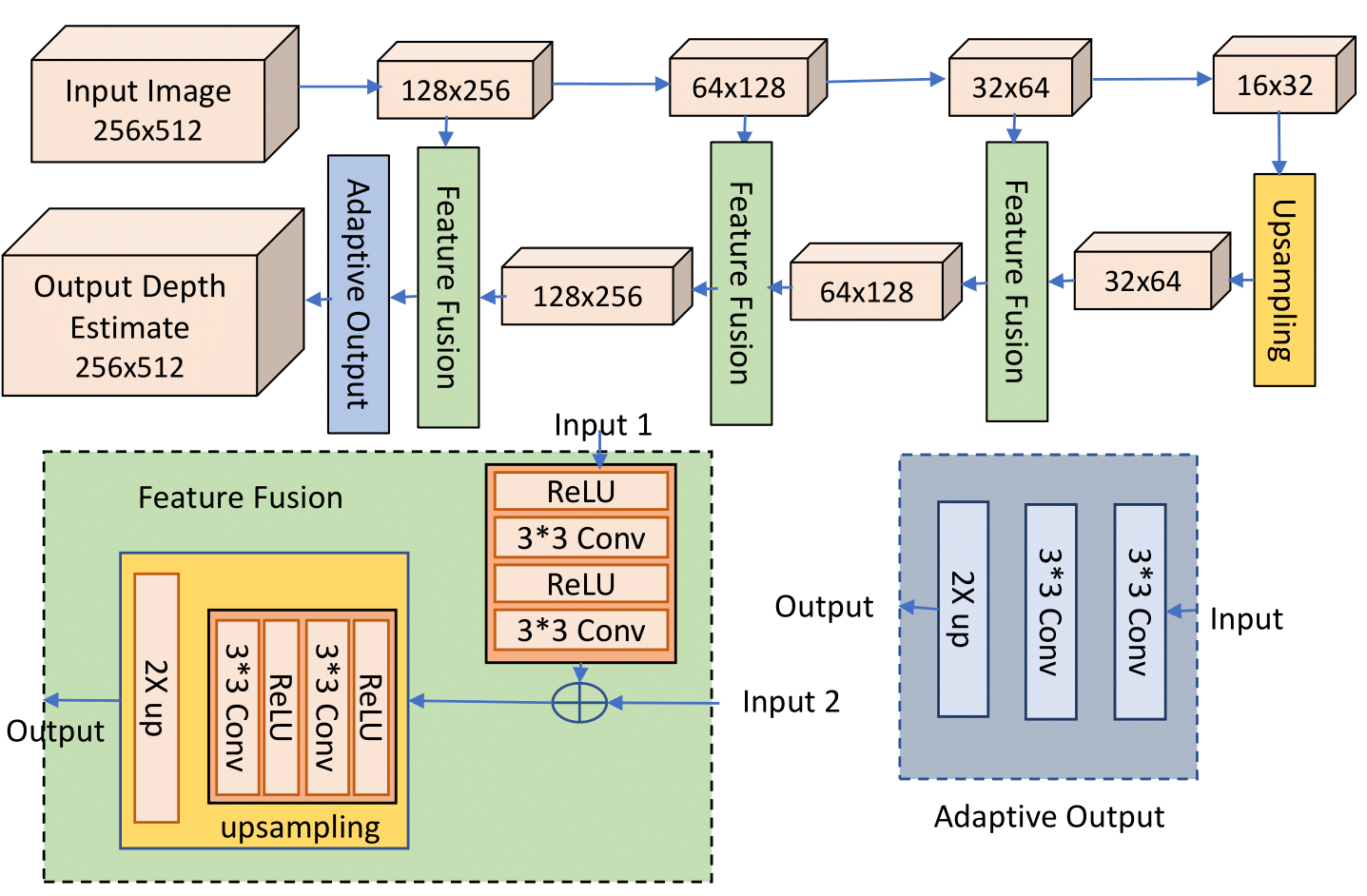}}
\caption{Schematic representation of feedforward ResNet Architecture: The network takes in a single view image and outputs the monocular depth estimate. The output is used as a depth clue in the Dual-Channel CNN architecture. }
\label{ResNetArc}
\end{figure}

\section{Proposed CNN Architecture}
\label{sec:ProposedCNNArchitecture}
We propose MEStereo-Du2CNN, a novel dual-channel CNN architecture to obtain robust depth estimates of a scene, given multi-exposure stereo images as input. The workflow of the proposed architecture is illustrated in Fig. \ref{MEStereoDu2CNNArc}. It can be described in two steps. The first step takes the stereo pairs at different exposure levels and  computes the disparity map using a dual-channel CNN structure (depicted in blue). The second step employs an exposure fusion technique to fuse the estimated disparity maps acquired from various exposure levels in the first step to obtain a refined disparity map. 

\subsection{Dual-channel architecture for disparity map prediction}

A novel and robust dual-channel CNN architecture predicts distinct disparity maps for stereo-image pairs at different exposure levels. This architecture has been extended from previously reported architecture by Anil et al. \cite{SDE-DualENet}. The components of proposed dual-channel CNN architecture are schematically described in Fig.~\ref{Du2CNNArc}. 

Most of the traditional stereo depth estimation algorithms work in four steps: Feature Extraction, Feature Matching, Disparity Estimation and Disparity Refinement. Feature matching works on the property that disparity between the same pixels on the left and the right viewpoints is indicative of the depth of that pixel. The pixels closer to the camera have a greater disparity between the viewpoints compared to the pixels further away. Stereo matching is performed by taking the patches centered around different points from the left and right images. These points are shifted in the $x$ direction by ‘$d$’ pixels. For each patch in the left image, ‘$d$’ is altered from $0-d_{max}$, and multiple patch pairs are obtained. For each patch pair, a similarity score is computed, and a cost volume is constructed. The construction of cost volume presents an additional variable  $d_{max}$, defined as the maximum disparity level up to which the stereo matching should be executed. 
The variable $d_{max}$ is a dataset-based preset parameter hard coded into the network.

We introduce a novel dual-channel CNN architecture that outperforms traditional stereo depth estimation algorithms. Traditional feature matching has been revamped to completely eliminate the need to construct the cost volume. We devise an alternative and more efficient method for utilising the information of disparity between the stereo pair for depth estimation. 

For every stereo pair at different exposure levels, we begin by using feed-forward ResNet-based multi-scale architecture of Ranftl et al. \cite{MiDaS2020} and Xian et al. \cite{Xian} to obtain the monocular depth clues of the left and the right images. The ResNet component as illustrated in Fig.~\ref{Du2CNNArc} of proposed MEStereo-Du2CNN architecture computes the monocular depth clues. It consists of a sequence of convolution and pooling operations as depicted in Fig.~\ref{ResNetArc}. To capture more contextual information, the resolution of input image is taken to be more than the output feature maps (32 times more). Post this, the multi-scale feature fusion operation in ResNet is employed to get a finer prediction out of the coarse prediction \cite{8100032}, \cite{8099589}. The computed multi-scale feature maps are progressively fused, by merging high level semantic features and low-level edge-sensitive features to further refine the prediction. Finally, an adaptive convolutional module adjusts the channels of feature maps and the final output. The ResNet architecture consists of multiple instances of upsampling and transitional convolutions, as illustrated in Fig.~\ref{ResNetArc}. 

The next step is to compute disparity map for the stereo pair using information obtained from monocular depth clues and exploiting disparity between the left and the right stereo views. The monocular depth clues from left and right views are fed into a novel dual-channel CNN network, which consists of a dual parallel encoder and single decoder network. The dual parallel encoder in our architecture uses different weights contrary to encoders with shared weight in traditional stereo depth estimation networks. To capture the disparities between left and right views at multiple resolution we employ a straightforward element-wise fusion method. At every resolution, the left and the right features are fused using an element-wise multiplication method. The result is then passed to the decoder using an element-wise addition method. The bilinear up-sampling operations maintain the resolution of the output by doubling the spatial resolution and halving the channel count of the feature map. The disparity map that is finally obtained from the decoder for given stereo-image pair has the exact resolution as those of the input images.

The dual parallel encoders are linked with feature fusion and back-propagation. As a result, the dual encoder weights are shifted to capture the stereo information. Hence, a feature point is aware of its adjacent points as well as captures disparity with respect to the other stereo view. During back-propagation, weights in the dual encoder are updated in a dependent fashion because of the element-wise multiplication of their features in the forward pass. This weight shift is functionally identical to the shifting of features for similarity calculation during cost volume construction. 

The convolutional blocks in the encoder side are based on EfficientNet architecture \cite{Tan2019EfficientNetRM}, which achieves better results compared to previous CNNs. Every block consists of a number of convolutional layers linked via skip connections. In terms of accuracy and efficiency, EfficientNet is better than previous ConvNets. This is because the baseline network of EfficentNet has been built by leveraging a multi-objective neural architecture search that optimizes both accuracy and FLOPS (floating point operations per second). The skip connections used for linking EfficientNets in the encoder are same as the ones present in ResNet block, except that instead of linking the layers with a higher channel count, the skip connections used in our architecture connect the lean channel layers. Thus, ensuring a lower computational cost and no loss of precision in the point-wise and channel-wise convolutions performed by layers in the EfficientNet.  

%Table 1
\begin{table*}[h!]
    \centering
    \caption{\centering{A Summary of three Middlebury stereo datasets. The "Number of Scenes" column counts only those scenes for which ground truth is available.}}
    \begin{tabular}{|c| c| c| c|} 
    \hline
    \textbf{Year} & \textbf{Number of Scenes}& \textbf{Resolution} & \textbf{Maximum Disparity}\\ [0.5ex]
 
    \hline
    Middlebury 2005 \cite{MB2005} & 6 & 1400 x 1100 & 230 \\ 
    %\hline
    Middlebury 2006 \cite{MB2006} & 21 & 1400 x 1100 & 230 \\
    %\hline
    Middlebury 2014 \cite{MB2014} & 23 & 3000 x 2000 & 800 \\
    \hline
    \end{tabular}
\label{table:MiddleBury_Data}
\end{table*}

\subsection{Fusion of Predicted Disparity Maps}

For a given scene, the second step merges disparity maps produced from stereo images at different exposure level. The procedure for fusing disparity maps is inspired from the work by Mertens et al. \cite{ExposureFusion}. The disparity maps are blended using the alpha masks, as inspired from Burt and Adelson \cite{1095851}. 
Given the input disparity maps, the weight map extraction method uses two quality measures, namely contrast and well-exposedness.

\textbf{Weight extraction using contrast}: We pass each disparity map through a Laplacian filter and consider the absolute value of filter response \cite{Malik:90}. This filter acts as an edge detector and assigns more weight to edges in the disparity map. A median blur filter further acts as a post processing method to smoothen out the discontinuities in detected edges, and thus preserve the sharp edges.

\textbf{Weight extraction using well-exposedness}:
In over-exposed and under-exposed parts of a captured image, the details of the scene are lost in the corresponding highlights and shadows. This leads to a poor depth estimation in such regions. Also, the regions of fewer details vary across images captured at different camera exposure levels. For example, a particular properly exposed region of an image has more details compared to the same corresponding region within another image of the same scene captured at a high shutter exposure camera setup. 

For a given exposure image $I_n$, the well-exposedness quality measure is extracted through a Gaussian curve applied on its grayscale as $\mathrm{exp(-({I_n}-0.5)^{2}/ 2{\sigma}^2)}$, where $\sigma = 0.2$ in our implementation. Each normalized pixel intensity of $I_n$ is weighted depending upon its closeness to 0.5. The aim is to allocate a higher weight to pixel intensities that are neither close to 0 (under-exposed) nor 1 (over-exposed). Hence, to favor pixels in well-exposed regions with intensity values close to 0.5. Higher weights are given to the pixels of properly exposed regions across differently exposed images of the same scene. The disparity maps corresponding to the stereo pairs at three exposure levels are blended using these weights.

\textbf{Weight refinement and fusion}: The information obtained from different quality measures is combined to form a refined weight map corresponding to each disparity map. We control the impact of each measure using corresponding ``weighting'' exponents $w_C$ and $w_E$. Refined weight map for $k^{th}$ disparity map at pixel position $(i,j)$ is given as

\begin{equation}
W_{i j, k}=\left(C_{i j, k}\right)^{\omega_{C}} \times\left(E_{i j, k}\right)^{\omega_{E}} 
\end{equation}

where, $0<k<N$ and $N$ is the number of obtained disparity estimates. Our architecture inputs three multi-exposed stereo pairs, which results in three intermediate disparity estimates; therefore, $N=3$.

If an exponent $w$ equals 0, the corresponding measure is not taken into account. Along every pixel, a weighted average is computed in order to fuse the $N$ disparity maps. To obtain a consistent result, we normalize values of $N$ weight maps such that at each pixel $(i, j)$ they sum to one, where $N$ is the total number of input disparity maps. The obtained weight maps are later combined to produce final fusion weights:

\begin{equation}
\hat{W}_{i j, k}=\left[\sum_{k^{\prime}=1}^{N} W_{i j, k^{\prime}}\right]^{-1} W_{i j, k}
\end{equation}

A straightforward way to obtain fused disparity map $R$ is by performing weighted blending of the input disparity maps as follows:

\begin{equation}
R_{i j}=\sum_{k=1}^{N} \hat{W}_{i j, k} D_{i j, k}
\end{equation}
where, $D_{k}$ represents $k^{th}$ input disparity map.

The problem with this approach is that disturbing seam emerges in the fused disparity map. Smoothing final weight maps with a Gaussian filter helps eliminate the abrupt weight map transitions, but results in unfavorable halos around the edges. We employ a method motivated by Burt and Adelson \cite{1095851} to solve this seam issue, where they use a pyramidal image decomposition to seamlessly merge two pictures directed by an alpha mask at varied resolutions. 

In our approach, $N$ final fusion weight maps, i.e., normalized weight maps, serve as alpha masks for the $N$ input disparity maps.
Each input disparity map is decomposed into $l$-levels of distinct resolutions using Laplacian pyramid (L). Similarly, the Gaussian pyramid (G) is utilized to decompose final fusion weights into $l$-levels of distinct resolutions. Let the $l^{th}$ level in a Laplacian pyramid decomposition of disparity map $D$ and Gaussian pyramid decomposition of final fusion weight map $\hat{W}$ be defined as $\mathrm{L}\{D\}^{l}$ and 
$\mathbf{G}\{\hat{W}\}^{l}$, respectively. The $N$ Laplacian pyramids $\mathrm{L}\{D\}$ are blended using Gaussian pyramid $\mathbf{G}\{\hat{W}\}$ to weight the $N$ disparity maps at each level of the pyramid as shown in the equation \ref{AlphaMask}, resulting in a reconstructed Laplacian pyramid decomposition $\mathrm{L}\{\hat{R}\}^{l}$, corresponding to the refined disparity map $\hat{R}$.

\begin{equation}
\mathbf{L}\{\hat{R}\}_{i j}^{l}=\sum_{k=1}^{N} \mathbf{G}\{\hat{W}\}_{i j, k}^{l} \mathbf{L}\{D\}_{i j, k}^{l}
\label{AlphaMask}
\end{equation}

The pyramid $\mathrm{L}\{\hat{R}\}^{l}$ is collapsed finally, to get the resulting refined disparity map $\hat{R}$. This method drastically improves the results of disparity fusion. The procedure is schematically shown in Fig.~\ref{MEStereoDu2CNNArc}. 

\section{Experiments and Results}
\label{sec:ExperimentsAndResults}
This section describes experimental results and performs a comparative analysis of the MEStereo-DU2CNN model with state-of-the-art CNN algorithms.

%COMPARATIVE ANALYSIS TABLES
%Table 2
\begin{table*}[h!]
    \centering
    
    \caption{\centering{Comparison of proposed MEStereo-DU2CNN architecture
    with different monocular \textit{(top)} and stereo \textit{(below)} based depth estimation algorithms on Scene flow test dataset. We train our model on Scene flow dataset. Best method per metric is highlighted in bold.}}
    
    \begin{tabular}{|c| c| c| c| c| c| c| c| c|} 
    \hline
    $\textbf{Method}$ & $\textit{$abs\_rel$}\downarrow$ & $\textit{$sq\_rel$}\downarrow$ & $\textit{$log\textsubscript{10}$}\downarrow$ & $RMSE\downarrow$ & $\sigma\textsubscript{1}\uparrow$ & $\sigma\textsubscript{2}\uparrow$ & $\sigma\textsubscript{3}\uparrow$ & $SSIM\uparrow$ \\ [0.5ex]
 
    \hline
    AdaBins \cite{Adabins2021} & 1.222 & 0.414 & 0.283 & 0.264 & 0.240 & 0.411 & 0.569 & 0.644 \\
    %\hline    
    CADepth \cite{CADepth2021} & 0.456 & 0.059 & 0.142 & 0.087 & 0.522 & 0.760 & 0.867 & 0.826 \\
    %\hline        
    DenseDepth \cite{DenseDepth2018} & 1.976 & 0.738 & 0.377 & 0.309 & 0.142 & 0.288 & 0.444 & 0.555 \\ 
    %\hline
    Depth Hints \cite{DepthHints2019} & 0.788 & 0.136 & 0.208 & 0.136 & 0.338 & 0.635 & 0.763 & 0.769 \\
    %\hline
    FCRN \cite{FCRN2016} & 1.143 & 0.288 & 0.280 & 0.219 & 0.208 & 0.394 & 0.599 & 0.658\\
    %\hline
    SerialUNet \cite{SerialUNet2020} & 0.861 & 0.165 & 0.216 & 0.151 & 0.344 & 0.576 & 0.735 & 0.713 \\
    %\hline
    MSDN \cite{MSDN2014} & 0.856 & 0.174 & 0.234 & 0.189 & 0.241 & 0.457 & 0.702 & 0.715 \\
    %\hline
    SIDE \cite{2019RevisitingSIDE} & 0.958 & 0.218 & 0.239 & 0.169 & 0.325 & 0.540 & 0.707 & 0.726 \\
    %\hline
    MiDaS \cite{MiDaS2020} & 0.338 & 0.031 & 0.146 & 0.072 & 0.550 & 0.782 & 0.879 & 0.840 \\
    %\hline

    \hline
    DeepROB \cite{DeepPruner2019} & 0.245 & 0.016 & 0.155 & 0.049 & 0.622 & 0.766 & 0.840 & 0.816 \\ 
    %\hline
    HSMNet \cite{HSMNet2019}& 0.485 & 0.106 & 0.286 & 0.164 & 0.383 & 0.517 & 0.600 & 0.685 \\
    %\hline
    PSMNet \cite{PSMNet2018} & 0.317 & 0.022 & 0.226 & 0.062 & 0.501 & 0.665 & 0.773 & 0.781\\
    %\hline
    STTR \cite{STTR2020} & 1.016 & 0.342 & 2.341 & 0.410 & 0.003 & 0.005 & 0.008 & 0.018 \\
    %\hline
    \textbf{MEStereo-Du2CNN (Ours)} & \textbf{0.193} & \textbf{0.010} & \textbf{0.109} & \textbf{0.038} & \textbf{0.663} & \textbf{0.827} & \textbf{0.895} & \textbf{0.864} \\  [1ex] 
    \hline
    
    \end{tabular}
\label{tab:SFSF}
\end{table*}

%Table 3
\begin{table*}[h!]
    \centering
    
    \caption{\centering{ Comparison of proposed MEStereo-DU2CNN architecture
    with different monocular \textit{(top)} and stereo \textit{(below)} based depth estimation algorithms on Middlebury 
    test dataset. We train our model on Middlebury dataset. Best method per metric is highlighted in bold.}}
    
    \begin{tabular}{|c| c| c| c|  c| c| c| c| c|} 
    \hline
    $\textbf{Method}$ & $\textit{$abs\_rel$}\downarrow$ & $\textit{$sq\_rel$}\downarrow$ & $\textit{log\textsubscript{10}}\downarrow$ & $RMSE\downarrow$ & $\sigma\textsubscript{1}\uparrow$ & $\sigma\textsubscript{2}\uparrow$ & $\sigma\textsubscript{3}\uparrow$ & $SSIM\uparrow$ \\ [0.5ex]
 
    \hline%\hline
    AdaBins \cite{Adabins2021} & 6.579 & 3.205 & 0.229 & 0.237 & 0.394 & 0.613 & 0.769 & 0.730 \\
    %\hline    
    CADepth \cite{CADepth2021} & 3.934 & 1.924 & 0.248 & 0.264 & 0.347 & 0.589 & 0.743 & 0.706 \\
    %\hline    
    DenseDepth \cite{DenseDepth2018} & 5.711 & 3.561 & 0.148 & 0.182 & 0.554 & 0.794 & 0.904 & 0.792 \\ 
    %\hline
    Depth Hints \cite{DepthHints2019} & 3.926 & 2.289 & 0.283 & 0.278 & 0.290 & 0.512 & 0.686 & 0.678 \\
    %\hline
    FCRN \cite{FCRN2016} & 6.186 & 4.185 & 0.190 & 0.246 & 0.442 & 0.678 & 0.818 & 0.737\\
    %\hline
    SerialUNet \cite{SerialUNet2020} & 5.613 & 3.685 & 0.175 & 0.226 & 0.465 & 0.710 & 0.854 & 0.730 \\
    %\hline
    MSDN \cite{MSDN2014} & 3.499 & 1.722 & 0.271 & 0.257 & 0.341 & 0.539 & 0.670 & 0.692 \\
    %\hline
    SIDE \cite{2019RevisitingSIDE} & 3.602 & 1.732 & 0.209 & 0.221 & 0.456 & 0.650 & 0.775 & 0.751 \\
    %\hline
    MiDaS \cite{MiDaS2020} & 2.656 & 1.025 & 0.335 & 0.244 & 0.346 & 0.518 & 0.637 & 0.700 \\
    \hline
    DeepROB \cite{DeepPruner2019} & 3.749 & 1.767 & 0.268 & 0.205 & 0.351 & 0.584 & 0.706 & 0.706 \\ 
    %\hline
    HSMNet \cite{HSMNet2019} & 3.361 & 1.436 & 0.249 & 0.187 & 0.453 & 0.611 & 0.698 & 0.732 \\
    %\hline
    PSMNet \cite{PSMNet2018} & 3.266 & 1.264 & 0.338 & 0.293 & 0.251 & 0.346 & 0.389 & 0.681\\
    %\hline
    STTR \cite{STTR2020}& \textbf{1.059} & 0.618 & 2.639 &  0.633 & 0.012 & 0.015 & 0.019 & 0.009 \\
    %\hline
    \textbf{MEStereo-Du2CNN (Ours)} & 1.549 & \textbf{0.549} & \textbf{0.072} & \textbf{0.079} & \textbf{0.846} & \textbf{0.939} & \textbf{0.970} & \textbf{0.884} \\  [1ex] 
    \hline
    
    \end{tabular}
\label{tab:MBMB}
\end{table*}

\subsection{Dataset}
The performance of proposed architecture is evaluated on three different data sets: Middlebury \cite{MB2005, MB2006,MB2014}, Scene flow \cite{SFFlying3D}, and multi-exposure natural scene stereo datasets \cite{HDR3DDataset}.

Middlebury dataset comprises high-resolution stereo sequence of static indoor scenes with an intricate geometry and pixel accurate ground truth disparity data acquired under controlled lighting conditions. Each scene in Middlebury 2005, 2006 and 2014 dataset was acquired under different lighting conditions, \textit{i.e.}, considering different illumination and exposure levels. A typical image pair of a scene captured under four lighting conditions and seven exposure settings making a total of 28 stereo pairs for the same scene. We train our model on Middlebury Stereo 2005, 2006, 2014 datasets \cite{MB2005}\cite{MB2006}\cite{MB2014}. The dataset used in our analysis consists of 50 RGB-D scenes as shown in Table \ref{table:MiddleBury_Data}. Each dataset scene consists of two views taken under different illuminations and with different exposures. The dataset has been provided in three different resolutions: full-size, half-size and third-size. We select third-size (width 443...463, height 370) from 2005 dataset. Likewise, we select third-size (width 413...465, height 370) from 2006 dataset. A standard train-test split of 90:10 is followed. while training our model. The training dataset has 847 stereo image pairs. The test dataset includes 27 scenes from 2005 and 2006 stereo data. For each test scene, we consider a single illumination and three exposure level stereo images, \textit{i.e.}, test dataset has 81 stereo pairs in total.

We also train the proposed model on synthetic scene flow driving data taken from  FlyingThings3D dataset \cite{SFFlying3D}. These dynamic scenes being quite natural are obtained from the first person perspective of the driver. It consists of about 4400 stereo scenes of trees, car models, roadside views, highly detailed objects such as trees and warehouse. For training the model we follow a typical 90:10 train-test split. We test on 440 image pairs from the FlyingThings3D scene flow data.

Our proposed model's performance is also evaluated on natural complex scenes. We use diverse stereoscopic 3D multi-exposure images database \cite{HDR3DDataset}, captured within the beautiful campus of Indian Institute of Technology Madras, India. The campus is a protected forest area, carved out of Guindy National Park. The campus is rich in flora and fauna and is a home of rare wildlife. The stereo database consists of complex natural scenes. the scenes contain dense trees, sky-scapes, endangered species of animals and birds, irregular reflecting surfaces, outdoor and indoor academic or residential area acquired under low-lit conditions. The scenes are complex for depth estimation task as dataset is rich in texture, color, details, exposure levels, depth structure, lightning conditions and object motions. The objects in some scenes have a slight motion between different exposure captures, such as forest trees swaying in the wind, rusting of the leaves, flowing water, etc. These scenes were acquired using ZED stereoscopic camera which has synchronized dual sensors separated at a distance of 12 cm from each other. The database consists of 38 different scenes captured in 2K (full HD) resolution at multiple exposures. Each image has a resolution of $2208 \times 1242$. We test our model using multi-exposure stereo pair sequences of all 38 scenes from the database.

\subsection{Implementation and Experimental settings} 
The model is implemented using PyTorch. The training and testing are executed on a single high-end HP OMEN X 15-DG0018TX 9th Gen i7-9750H Gaming laptop, 16 GB RAM, RTX 2080 8 GB Graphics and Windows 10 operating system.

We train proposed model on scene flow dataset for 10 epochs and 495 iterations per epoch. Training on scene flow takes about 11 hours with an inference time of 140 milliseconds for each stereo image pair. The model is also trained on the Middlebury dataset for 70 epochs and 96 iterations per epoch. Training on Middlebury dataset takes around 13 hours. The testing time of the model is around 26 milliseconds for a stereo image pair.

\subsection{Comparative Analysis}
Our proposed model is compared with the latest state-of-the-art monocular and stereo based depth estimation algorithms. We select nine monocular depth estimation algorithms:  AdaBins \cite{Adabins2021}, CADepth \cite{CADepth2021}, Depth Hints \cite{DepthHints2019}, DenseDepth \cite{DenseDepth2018}, FCRN \cite{FCRN2016}, SerialUNet \cite{SerialUNet2020}, SIDE \cite{2019RevisitingSIDE}, MSDN \cite{MSDN2014}, MiDaS \cite{MiDaS2020}; and four stereo depth estimation algorithms: DeepPruner \cite{DeepPruner2019}, HSMNet  \cite{HSMNet2019}, PSMNet \cite{PSMNet2018} and STTR \cite{STTR2020}. We use publicly available pre-trained models for evaluating the comparison methods.

The results of monocular depth estimation methods are calculated considering  left and right view of stereo pair individually. To obtain the corresponding monocular depth map, one view is processed at a time. However, the left and right views are taken as input simultaneously for obtaining depth map results using stereo algorithms.

We use standard error metrics for quantitative analysis: Absolute relative error ($abs rel$), Squared relative error ($sq rel$), Root mean square error (RMSE), Average log error ($log_{10}$), threshold accuracy ($\sigma_i$) and  perception-based Structural Similarity Index Metric (SSIM) \cite{MiDaS2020}\cite{3DGBUNet2020}. Given a predicted depth image and its corresponding ground truth, the different error metrics are calculated as follows:

\textit{Absolute relative error} :
\begin{equation}
\textit{abs\_rel} = \frac{1}{|T|} \sum_{p\epsilon T} \frac{|y_{p} - y_{p}^{*}|}{y_{p}^{*}}
\end{equation}

\textit{Squared relative error} :
\begin{equation}
\textit{sq\_rel} = \frac{1}{|T|} \sum_{p\epsilon T} \frac{||y_{p} - y_{p}^{*}||^{2}}{y_{p}^{*}}
\end{equation}

\textit{Root mean square error} :
\begin{equation}
\textit{RMSE} = \sqrt{\frac{1}{|T|} \sum_{p\epsilon T} ||y_{p} - y_{p}^{*}||^{2}}
\end{equation}

\textit{Average log error} :
\begin{equation}
\textit{$log_{10}$} = \sqrt{\frac{1}{|T|} \sum_{p\epsilon T} ||log y_{p} - log y_{p}^{*}||^{2}}
\end{equation}

\textit{Threshold accuracy} : percentage of $y_p$ such that
\begin{equation}
max(\frac{y_{p}}{y_{p}^*},\frac{y_{p}^*}{y_{p}}) = \sigma_i < thres
\end{equation}
for $thres = 1.25, 1.25^2, 1.25^3$.

%%%%%%% FIGURES

%FIGURE 4
\begin{figure*}[h!]
{\includegraphics[height=0.60in,width=0.99in]{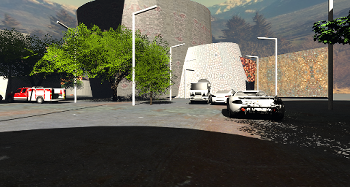}}\hfill
{\includegraphics[height=0.60in,width=0.99in]{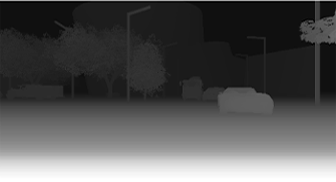}}\hfill
{\includegraphics[height=0.60in,width=0.99in]{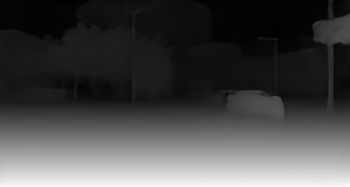}}\hfill
{\includegraphics[height=0.60in,width=0.99in]{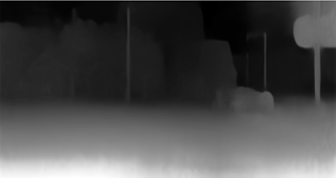}}\hfill
{\includegraphics[height=0.60in,width=0.99in]{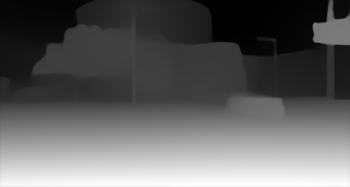}}\hfill
{\includegraphics[height=0.60in,width=0.99in]{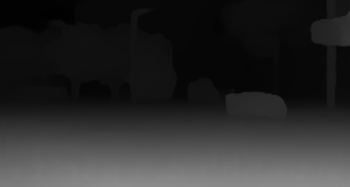}}\hfill
{\includegraphics[height=0.60in,width=0.99in]{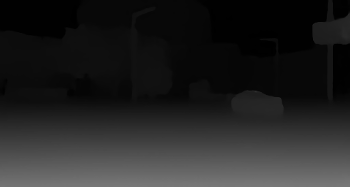}}\hfill

{\includegraphics[height=0.60in,width=0.99in]{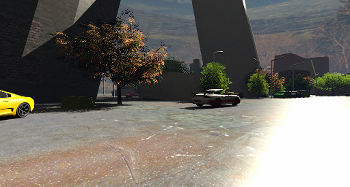}}\hfill
{\includegraphics[height=0.60in,width=0.99in]{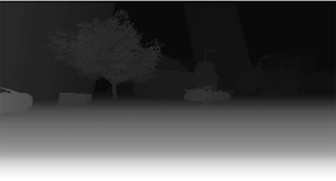}}\hfill
{\includegraphics[height=0.60in,width=0.99in]{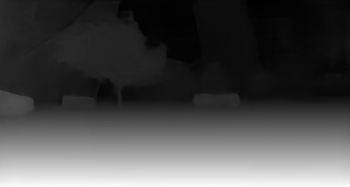}}\hfill
{\includegraphics[height=0.60in,width=0.99in]{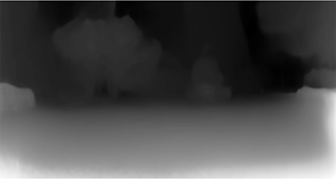}}\hfill
{\includegraphics[height=0.60in,width=0.99in]{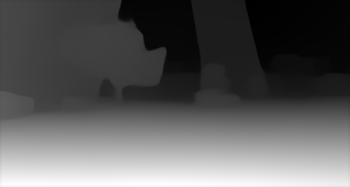}}\hfill
{\includegraphics[height=0.60in,width=0.99in]{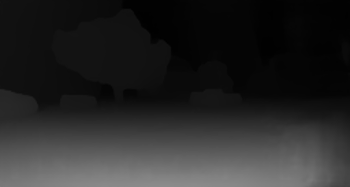}}\hfill
{\includegraphics[height=0.60in,width=0.99in]{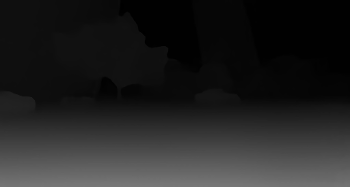}}\hfill

\vspace*{-3.5mm}
{\subfloat[Left image]{\includegraphics[height=0.60in,width=0.99in]{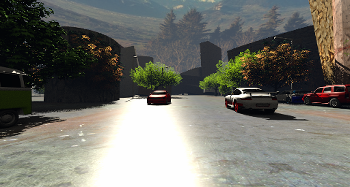}}}\hfill
{\subfloat[Ground Truth]{\includegraphics[height=0.60in,width=0.99in]{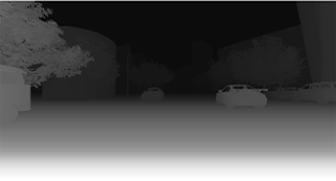}}}\hfill
{\subfloat[Proposed]{\includegraphics[height=0.60in,width=0.99in]{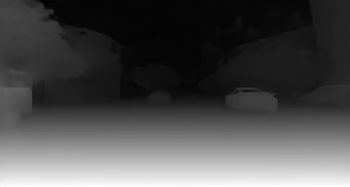}}}\hfill
{\subfloat[Depth Hints\cite{DepthHints2019}]{\includegraphics[height=0.60in,width=0.99in]{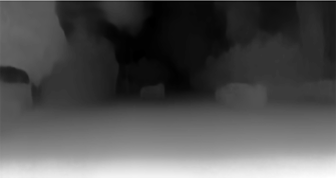}}}\hfill
{\subfloat[MiDaS\cite{MiDaS2020}]{\includegraphics[height=0.60in,width=0.99in]{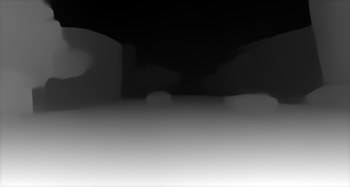}}}\hfill
{\subfloat[DeepROB\cite{DeepPruner2019}]{\includegraphics[height=0.60in,width=0.99in]{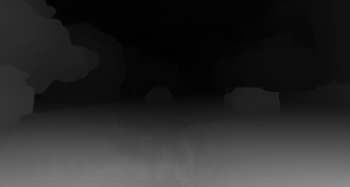}}}\hfill
{\subfloat[PSMNet\cite{PSMNet2018}]{\includegraphics[height=0.60in,width=0.99in]{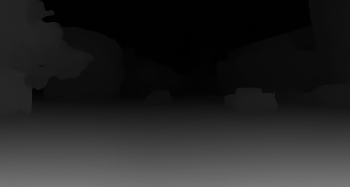}}}\hfill

\caption{Visual comparison results of proposed MEStereo-DU2CNN architecture trained on Scene flow dataset. We compare with state-of-the-art monocular \textit{(d), (e)} and stereo \textit{(f), (g)} depth estimation algorithms.}
\label{FIG_SF}
\end{figure*}

%Figure 5
\begin{figure*}[h!]
{\includegraphics[height=0.75 in,width=0.99in]{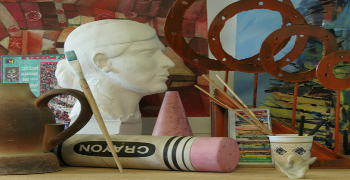}}\hfill
{\includegraphics[height=0.75 in,width=0.99in]{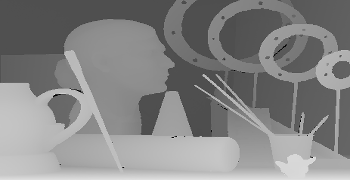}}\hfill
{\includegraphics[height=0.75 in,width=0.99in]{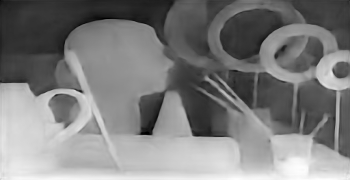}}\hfill
{\includegraphics[height=0.75 in,width=0.99in]{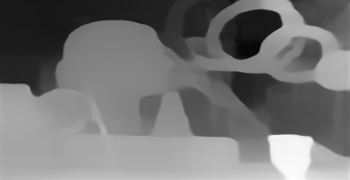}}\hfill
{\includegraphics[height=0.75 in,width=0.99in]{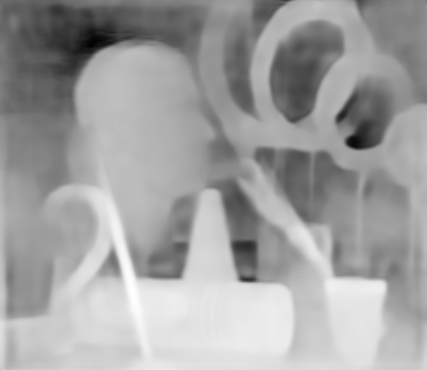}}\hfill
{\includegraphics[height=0.75 in,width=0.99in]{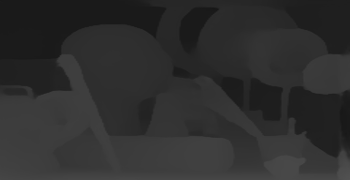}}\hfill
{\includegraphics [height=0.75 in,width=0.99in]{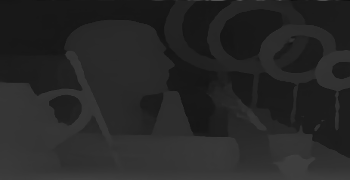}}\hfill

{\includegraphics[height=0.75 in,width=0.99in]{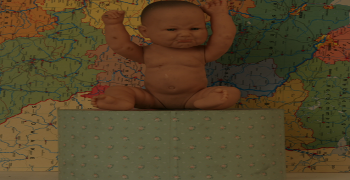}}\hfill
{\includegraphics[height=0.75 in,width=0.99in]{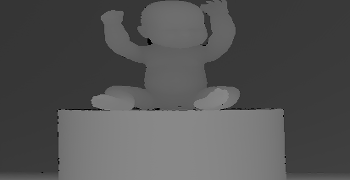}}\hfill
{\includegraphics[height=0.75 in,width=0.99in]{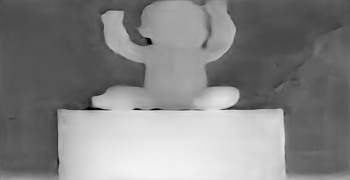}}\hfill
{\includegraphics[height=0.75 in,width=0.99in]{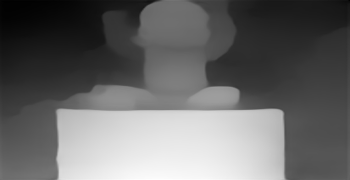}}\hfill
{\includegraphics[height=0.75 in,width=0.99in]{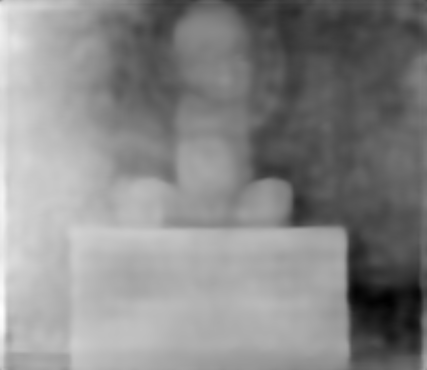}}\hfill
{\includegraphics[height=0.75 in,width=0.99in]{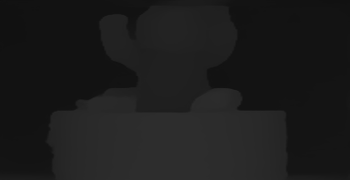}}\hfill
{\includegraphics [height=0.75 in,width=0.99in]{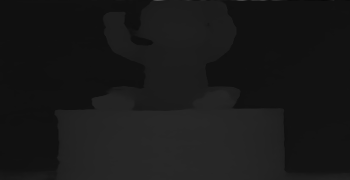}}\hfill

\vspace*{-3.5mm}
{\subfloat[Left image]{\includegraphics[height=0.75 in,width=0.99in]{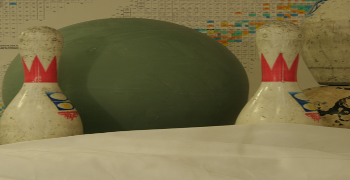}}}\hfill
{\subfloat[Ground Truth]{\includegraphics[height=0.75 in,width=0.99in]{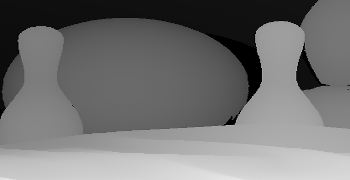}}}\hfill
{\subfloat[Proposed]{\includegraphics[height=0.75 in,width=0.99in]{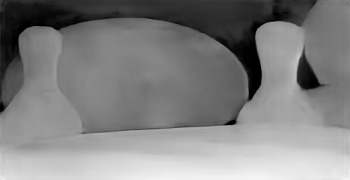}}}\hfill
{\subfloat[MiDaS\cite{MiDaS2020}]{\includegraphics[height=0.75 in,width=0.99in]{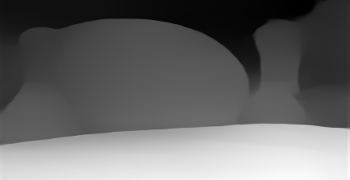}}}\hfill
{\subfloat[AdaBins\cite{Adabins2021}]{\includegraphics[height=0.75 in,width=0.99in]{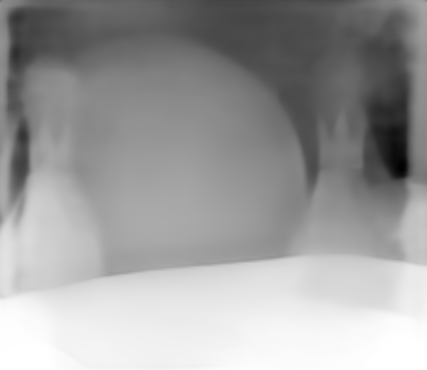}}}\hfill
{\subfloat[DeepROB\cite{DeepPruner2019}]{\includegraphics[height=0.75 in,width=0.99in]{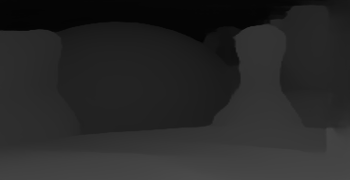}}}\hfill
{\subfloat[PSMNet\cite{PSMNet2018}]{\includegraphics[height=0.75 in,width=0.99in]{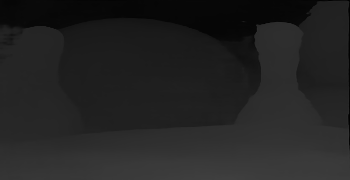}}}\hfill

\caption{Visual comparison results of proposed MEStereo-DU2CNN architecture trained on Middlebury dataset. We compare
with state-of-the-art monocular \textit{(d), (e)} and stereo \textit{(f), (g)} depth estimation algorithms.}
\label{FIG_MB}
\end{figure*}

%FIGURE 6
\begin{figure*}[h!]
{\includegraphics[height=0.70in ,width=0.99in]{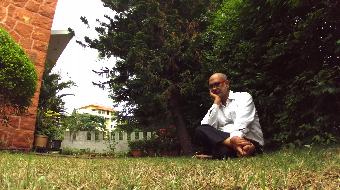}}\hfill
{\includegraphics[height=0.70in ,width=0.99in]{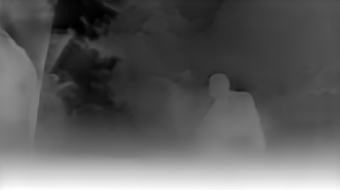}}\hfill
{\includegraphics[height=0.70in
,width=0.99in]{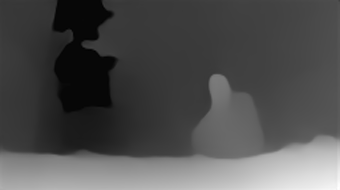}}\hfill
{\includegraphics [height=0.70in
,width=0.99in]{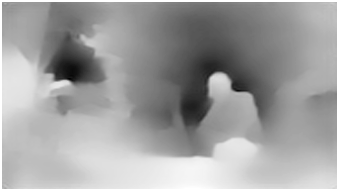}}\hfill
{\includegraphics[height=0.70in ,width=0.99in]{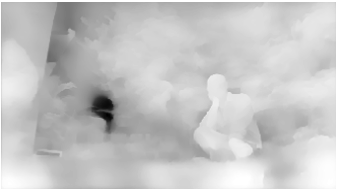}}\hfill
{\includegraphics[height=0.70in ,width=0.99in]{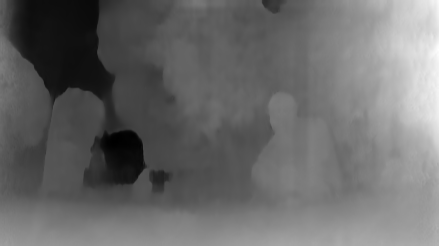}}\hfill
{\includegraphics [height=0.70in
,width=0.99in]{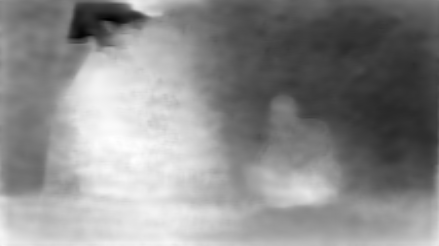}}\hfill

\vspace*{-3.5mm}

{\subfloat[Left image]{\includegraphics[height=0.70in ,width=0.99in]{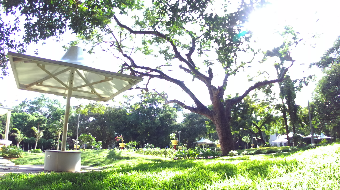}}}\hfill
{\subfloat[Proposed]{\includegraphics[height=0.70in ,width=0.99in]{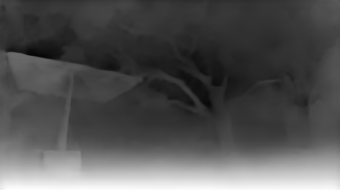}}}\hfill
{\subfloat[MiDaS\cite{MiDaS2020}]{\includegraphics[height=0.70in ,width=0.99in]{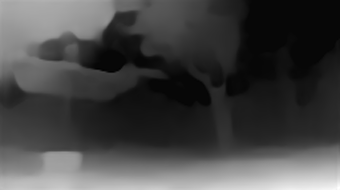}}}\hfill
{\subfloat[SIDE\cite{2019RevisitingSIDE}]{\includegraphics[height=0.70in ,width=0.99in]{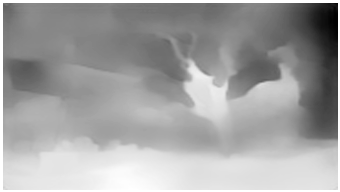}}}\hfill
{\subfloat[DenseDepth\cite{DenseDepth2018}]{\includegraphics[height=0.70in ,width=0.99in]{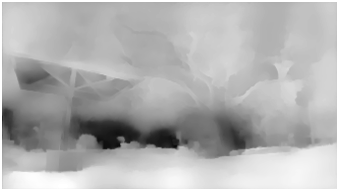}}}\hfill
{\subfloat[CADepth\cite{CADepth2021}]{\includegraphics[height=0.70in ,width=0.99in]{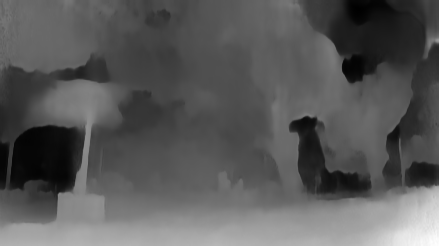}}}\hfill
{\subfloat[AdaBins\cite{Adabins2021}]{\includegraphics[height=0.70in ,width=0.99in]{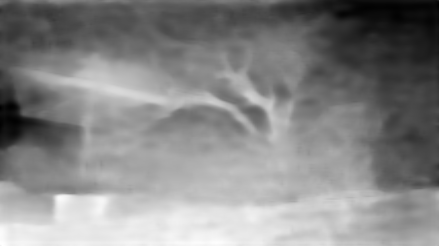}}}\hfill
\caption{Visual comparison results of proposed MEStereo-DU2CNN
architecture  trained  on Scene flow dataset and tested on natural stereoscopic 3D multi-exposure scenes. We compare with state-of-the-art depth estimation algorithms.}
\label{FIG_ZED}
\end{figure*}

Here, $y_p^*$ denotes the predicted value of depth at pixel $p$, $y_p$ denotes the ground truth value of depth at pixel $p$, $T$ denotes the total number of pixels for which there exists both valid ground truth and predicted depth.

Lower values are better for $absrel$, $sqrel$, $log_{10}$ and RMSE. Higher values indicate better quality for $\sigma_1$, $\sigma_2$, $\sigma_3$ and SSIM measures. We compare performance of our MEStereo-DU2CNN architecture against the existing state-of-the-art monocular and stereo based depth estimation methods \cite{Adabins2021, CADepth2021, DenseDepth2018,DepthHints2019,FCRN2016,SerialUNet2020,MSDN2014,2019RevisitingSIDE,MiDaS2020,DeepPruner2019,HSMNet2019,PSMNet2018,STTR2020}. We use available pre-trained models of baseline methods. The results are presented on Scene flow, Middlebury and complex natural scenes.  

\textbf{Evaluation on Scene flow}:
Our model gives encouraging results on Scene flow dataset with higher quality depth maps. Table \ref{tab:SFSF} shows the quantitative analysis of MEStereo-DU2UCNN architecture on Scene flow compared with other baseline methods. Comparative visual results are shown in Figure \ref{FIG_SF}. We choose three scenes which include reflection and shadows on tree, car and buildings. These scenes have black walls and large shadow areas. There are practically no visible textural cues to aid in the identification of corresponding matching points. Also, reflective glass and road surface are ill-posed areas. Our model outperforms other methods both quantitatively and qualitatively and has more robust depth estimation results particularly in the regions of car windows and wall.

\textbf{Evaluation on Middlebury}:
Quantitative analysis of MEStereo-DU2UCNN architecture on Middlebury dataset with respect to other baseline methods is shown in Table \ref{tab:MBMB}. Our model significantly outperforms state-of-the-art monocular and stereo based depth estimation methods by a good margin across the given metrics. For qualitative comparison, we choose $Art$ from Middlebury 2005 dataset \cite{MB2005}, $Baby1$ and $Bowling1$ from Middlebury 2006 dataset \cite{MB2006}. As shown in Figure \ref{FIG_MB}, our method produces smooth depth planes and sharp estimation on object boundaries. Also, MEStereo-DU2CNN is able to capture large disparities in Middlebury dataset.

%The visual comparison results of proposed model with other monocular and stereo methods on scene flow, Middlebury and complex natural scenes are shown in Figure \ref{FIG_SF_MONO}-\ref{FIG_ZED_Stereo}.

\textbf{Evaluation on complex natural scenes}:
The task of estimating depth in a natural scene characterised by complex motions, changes in lighting, and illumination is challenging. To show effectiveness of our approach, we perform visual comparison with other methods on complex natural scenes, as depicted in Figure \ref{FIG_ZED}. We use Scene flow trained MEStereo-DU2CNN architecture for this task. Our proposed model outperforms other algorithms. The quantitative analysis for complex natural scenes is not performed due to the lack of ground truth data.

The depth can be obtained from disparity map as given in the equation below:

\begin{equation}
\text{depth} = \frac{\text{baseline} \times \text{focal length in pixels}}{\text{disparity}} 
\label{equation10}
\end{equation}

where, baseline is the distance between the left and the right cameras. The unit of depth is the same as that of baseline.

The parameters for acquiring depth from disparity for different datasets are provided on their respective websites, Middlebury 2005 \cite{MB2005}, Middlebury 2006 \cite{MB2006}, Middlebury 2014 \cite{MB2014}, Scene flow \cite{SFFlying3D} and stereoscopic 3D multi-exposure images database of natural scenes \cite{HDR3DDataset}.

\section{Conclusion}
\label{sec:Conclusion}
We have proposed a novel end-to-end CNN architecture for robust depth prediction using multi-exposed stereo image sequences. The stereo depth estimation component used in our architecture simultaneously uses a mono-to-stereo dual-transfer learning approach along with the replacement of conventional cost volume construction. Encoders with shared weight used in traditional stereo matching methods are replaced by a novel ResNet based dual-encoder single-decoder framework with different weights. EfficientNet based blocks are used in  convolutions layers of the dual encoders. The dual encoder weights are shifted rather than feature maps shift at various disparity levels, thereby avoiding the need to specify a scene's disparity range. Therefore, the proposed method addresses major limitations of the current stereo depth estimation algorithms, which do not give satisfactory results in low-texture over- or under-exposed image regions, natural lighting conditions and detail structures. The disparity maps obtained at different exposure stereo pairs are fused to refine disparity predictions further. 

The proposed model completely bypasses the need for tone-mapped SHDR images for stereo matching. Also, it avoids complicated process to generate depth maps from floating point values stored in HDR data. Instead, we aim to develop a model that completely eliminates the necessity of having expensive HDR stereo inputs and replace them with affordable multi-exposure SDR images by effectively handling dynamic range locally or globally for predicting depth in practical 3D applications. We want to expand the proposed dual-parallel CNN for stereo-to-multiview rendering system for view synthesis and VR, 3D display, free viewpoint video applications.

\bibliographystyle{plain}
\bibliography{root}

\begin{thebibliography}{10}

\bibitem{HDRImaging2015}
Nayana A. and Anoop K.~Johnson.
\newblock High dynamic range imaging- a review.
\newblock {\em International Journal of Image Processing (IJIP)}, 9(4), 2015.

\bibitem{Akhavan2015BackwardCH}
Tara Akhavan and Hannes Kaufmann.
\newblock Backward compatible hdr stereo matching: a hybrid tone-mapping-based
  framework.
\newblock {\em EURASIP JIVP}, 2015:1--12, 2015.

\bibitem{Akhavan2013AFF}
Tara Akhavan, Hyunjin Yoo, and M.~Gelautz.
\newblock A framework for hdr stereo matching using multi-exposed images.
\newblock In {\em HDRi2013}, pages 1--4, 2013.

\bibitem{DenseDepth2018}
Ibraheem Alhashim and Peter Wonka.
\newblock High quality monocular depth estimation via transfer learning.
\newblock {\em arXiv e-prints}, abs/1812.11941, 2018.

\bibitem{SDE-DualENet}
Rithvik Anil, Mansi Sharma, and Rohit Choudhary.
\newblock Sde-dualenet: A novel dual efficient convolutional neural network for
  robust stereo depth estimation.
\newblock In {\em 2021 International Conference on Visual Communications and
  Image Processing (VCIP)}, pages 1--5, 2021.

\bibitem{1095851}
P.~Burt and E.~Adelson.
\newblock The laplacian pyramid as a compact image code.
\newblock {\em IEEE Trans Commun}, 31(4):532--540, 1983.

\bibitem{SerialUNet2020}
Kyle~J. Cantrell., Craig~D. Miller., and Carlos~W. Morato.
\newblock Practical depth estimation with image segmentation and serial u-nets.
\newblock In {\em VEHITS}, volume~I, pages 406--414. INSTICC, SciTePress, 2020.

\bibitem{PSMNet2018}
Jia-Ren Chang and Yong-Sheng Chen.
\newblock Pyramid stereo matching network.
\newblock In {\em IEEE CVPR}, pages 5410--5418, 2018.

\bibitem{Chari2020OptimalHA}
P.~Chari, Anil~Kumar Vadathya, and K.~Mitra.
\newblock Optimal hdr and depth from dual cameras.
\newblock {\em ArXiv}, abs/2003.05907, 2020.

\bibitem{DeepPruner2019}
Shivam Duggal, Shenlong Wang, Wei-Chiu Ma, Rui Hu, and Raquel Urtasun.
\newblock Deeppruner: Learning efficient stereo matching via differentiable
  patchmatch.
\newblock In {\em ICCV}, 2019.

\bibitem{MSDN2014}
David Eigen, Christian Puhrsch, and Rob Fergus.
\newblock Depth map prediction from a single image using a multi-scale deep
  network.
\newblock In {\em Adv Neural Inf Process Syst}, volume~27. Curran Associates,
  Inc., 2014.

\bibitem{Eilertsen2017HDRIReSingleEx}
Gabriel Eilertsen, Joel Kronander, Gyorgy Denes, Rafa\l{}~K. Mantiuk, and Jonas
  Unger.
\newblock Hdr image reconstruction from a single exposure using deep cnns.
\newblock {\em ACM Trans. Graph.}, 36(6), nov 2017.

\bibitem{Adabins2021}
Shariq Farooq~Bhat, Ibraheem Alhashim, and Peter Wonka.
\newblock Adabins: Depth estimation using adaptive bins.
\newblock In {\em 2021 IEEE/CVF Conference on Computer Vision and Pattern
  Recognition (CVPR)}, pages 4008--4017, 2021.

\bibitem{SIuAGNs2018}
Zhixiang Hao, Yu~Li, Shaodi You, and Feng Lu.
\newblock Detail preserving depth estimation from a single image using
  attention guided networks.
\newblock In {\em 2018 International Conference on 3D Vision (3DV)}, pages
  304--313, 2018.

\bibitem{Hasinoff2010}
Samuel~W. Hasinoff, Frédo Durand, and William~T. Freeman.
\newblock Noise-optimal capture for high dynamic range photography.
\newblock In {\em IEEE CVPR}, pages 553--560, 2010.

\bibitem{SMsemiglobal2008}
Heiko Hirschmuller.
\newblock Stereo processing by semiglobal matching and mutual information.
\newblock {\em IEEE Transactions on Pattern Analysis and Machine Intelligence},
  30(2):328--341, 2008.

\bibitem{MB2006}
Heiko Hirschmuller and Daniel Scharstein.
\newblock Evaluation of cost functions for stereo matching.
\newblock In {\em IEEE CVPR}, pages 1--8, 2007,
  \url{https://vision.middlebury.edu/stereo/data/scenes2006/}.

\bibitem{2019RevisitingSIDE}
Junjie Hu, Mete Ozay, Yan Zhang, and Takayuki Okatani.
\newblock Revisiting single image depth estimation: Toward higher resolution
  maps with accurate object boundaries.
\newblock In {\em IEEE WACV}, 2019.

\bibitem{Im:AutoExpBracketing}
Sunghoon Im, Hae-Gon Jeon, and In~So Kweon.
\newblock Robust depth estimation using auto-exposure bracketing.
\newblock {\em IEEE Transactions on Image Processing}, 28(5):2451--2464, 2019.

\bibitem{Kalantari2017DHDRIDynScenes}
Nima~Khademi Kalantari and Ravi Ramamoorthi.
\newblock Deep high dynamic range imaging of dynamic scenes.
\newblock {\em ACM Trans. Graph.}, 36(4), jul 2017.

\bibitem{KendallDeepStereoRegression2017}
Alex Kendall, Hayk Martirosyan, Saumitro Dasgupta, Peter Henry, Ryan Kennedy,
  Abraham Bachrach, and Adam Bry.
\newblock End-to-end learning of geometry and context for deep stereo
  regression.
\newblock In {\em 2017 IEEE International Conference on Computer Vision
  (ICCV)}, pages 66--75, 2017.

\bibitem{FCRN2016}
Iro Laina, Christian Rupprecht, Vasileios Belagiannis, Federico Tombari, and
  Nassir Navab.
\newblock Deeper depth prediction with fully convolutional residual networks.
\newblock In {\em 3DV}, pages 239--248, 2016.

\bibitem{STTR2020}
Zhaoshuo Li, Xingtong Liu, Nathan Drenkow, Andy Ding, Francis~X Creighton,
  Russell~H Taylor, and Mathias Unberath.
\newblock Revisiting stereo depth estimation from a sequence-to-sequence
  perspective with transformers.
\newblock {\em arXiv:2011.02910}, 2020.

\bibitem{LiangFeatureConstancy2018}
Zhengfa Liang, Yiliu Feng, Yulan Guo, Hengzhu Liu, Wei Chen, Linbo Qiao,
  Li~Zhou, and Jianfeng Zhang.
\newblock Learning for disparity estimation through feature constancy.
\newblock In {\em 2018 IEEE/CVF Conference on Computer Vision and Pattern
  Recognition}, pages 2811--2820, 2018.

\bibitem{8100032}
Guosheng Lin, Anton Milan, Chunhua Shen, and Ian Reid.
\newblock Refinenet: Multi-path refinement networks for high-resolution
  semantic segmentation.
\newblock In {\em IEEE CVPR}, pages 5168--5177, 2017.

\bibitem{HueiYung2016}
Huei-Yung Lin and Chung-Chieh Kao.
\newblock Stereo matching techniques for high dynamic range image pairs.
\newblock In {\em Image and Video Technology}, pages 605--616, 2016.

\bibitem{8099589}
Tsung-Yi Lin, Piotr Dollár, Ross Girshick, Kaiming He, Bharath Hariharan, and
  Serge Belongie.
\newblock Feature pyramid networks for object detection.
\newblock In {\em IEEE CVPR}, pages 936--944, 2017.

\bibitem{Liu2020SImageHDR}
Yu-Lun Liu, Wei-Sheng Lai, Yu-Sheng Chen, Yi-Lung Kao, Ming-Hsuan Yang, Yung-Yu
  Chuang, and Jia-Bin Huang.
\newblock Single-image hdr reconstruction by learning to reverse the camera
  pipeline.
\newblock In {\em 2020 IEEE/CVF Conference on Computer Vision and Pattern
  Recognition (CVPR)}, pages 1648--1657, 2020.

\bibitem{Malik:90}
Jitendra Malik and Pietro Perona.
\newblock Preattentive texture discrimination with early vision mechanisms.
\newblock {\em Journal of the Optical Society of America. A}, 7(5):923--932,
  May 1990.

\bibitem{SFFlying3D}
Nikolaus Mayer, Eddy Ilg, Philip Häusser, Philipp Fischer, Daniel Cremers,
  Alexey Dosovitskiy, and Thomas Brox.
\newblock A large dataset to train convolutional networks for disparity,
  optical flow, and scene flow estimation.
\newblock In {\em 2016 IEEE Conference on Computer Vision and Pattern
  Recognition (CVPR)}, pages 4040--4048, 2016,
  \url{https://lmb.informatik.uni-freiburg.de/resources/datasets/SceneFlowDatasets.en.html}.

\bibitem{ExposureFusion}
Tom Mertens, Jan Kautz, and Frank Van~Reeth.
\newblock Exposure fusion.
\newblock In {\em 15th Pacific Conference on Computer Graphics and Applications
  (PG'07)}, pages 382--390, 2007.

\bibitem{Mozerov2015SM2StepEnergyMin}
Mikhail~G. Mozerov and Joost van~de Weijer.
\newblock Accurate stereo matching by two-step energy minimization.
\newblock {\em IEEE Transactions on Image Processing}, 24(3):1153--1163, 2015.

\bibitem{Ning2018LearningInvTM}
Shiyu Ning, Hongteng Xu, Li~Song, Rong Xie, and Wenjun Zhang.
\newblock Learning an inverse tone mapping network with a generative
  adversarial regularizer.
\newblock In {\em 2018 IEEE International Conference on Acoustics, Speech and
  Signal Processing (ICASSP)}, pages 1383--1387, 2018.

\bibitem{IntraInterDP1985}
Yuichi Ohta and Takeo Kanade.
\newblock Stereo by intra- and inter-scanline search using dynamic programming.
\newblock {\em IEEE Transactions on Pattern Analysis and Machine Intelligence},
  PAMI-7(2):139--154, 1985.

\bibitem{MiDaS2020}
Rene Ranftl, Katrin Lasinger, David Hafner, Konrad Schindler, and Vladlen
  Koltun.
\newblock Towards robust monocular depth estimation: Mixing datasets for
  zero-shot cross-dataset transfer.
\newblock {\em IEEE TPAMI}, pages 1--1, 2020.

\bibitem{TaxEvalStereoCorres2001}
D.~Scharstein, R.~Szeliski, and R.~Zabih.
\newblock A taxonomy and evaluation of dense two-frame stereo correspondence
  algorithms.
\newblock In {\em Proceedings IEEE Workshop on Stereo and Multi-Baseline Vision
  (SMBV 2001)}, pages 131--140, 2001.

\bibitem{MB2014}
Daniel Scharstein, Heiko Hirschm{\"u}ller, York Kitajima, Greg Krathwohl, Nera
  Ne{\v{s}}i{\'{c}}, Xi~Wang, and Porter Westling.
\newblock High-resolution stereo datasets with subpixel-accurate ground truth.
\newblock In Xiaoyi Jiang, Joachim Hornegger, and Reinhard Koch, editors, {\em
  Pattern Recognition}, pages 31--42, 2014,
  \url{https://vision.middlebury.edu/stereo/data/scenes2014}.

\bibitem{MB2005}
Daniel Scharstein and Chris Pal.
\newblock Learning conditional random fields for stereo.
\newblock In {\em IEEE CVPR}, pages 1--8, 2007,
  \url{https://vision.middlebury.edu/stereo/data/scenes2005/}.

\bibitem{3DGBUNet2020}
Mansi Sharma, Abheesht Sharma, Kadvekar~Rohit Tushar, and Avinash Panneer.
\newblock A novel 3d-unet deep learning framework based on high-dimensional
  bilateral grid for edge consistent single image depth estimation.
\newblock In {\em IC3D}, pages 01--08, 2020.

\bibitem{Tan2019EfficientNetRM}
Mingxing Tan and Quoc~V. Le.
\newblock Efficientnet: Rethinking model scaling for convolutional neural
  networks.
\newblock {\em ArXiv}, abs/1905.11946, 2019.

\bibitem{HDR3DDataset}
Aditya Wadaskar, Mansi Sharma, and Rohan Lal.
\newblock A rich stereoscopic 3d high dynamic range image amp; video database
  of natural scenes.
\newblock In {\em IC3D}, pages 1--8, 2019,
  \url{https://ieeexplore.ieee.org/document/8975903}.

\bibitem{Wang2021DLforHDRI}
Lin Wang and Kuk-Jin Yoon.
\newblock Deep learning for hdr imaging: State-of-the-art and future trends.
\newblock {\em IEEE Transactions on Pattern Analysis and Machine Intelligence},
  pages 1--1, 2021.

\bibitem{DepthHints2019}
Jamie Watson, Michael Firman, Gabriel~J. Brostow, and Daniyar Turmukhambetov.
\newblock Self-supervised monocular depth hints.
\newblock In {\em IEEE ICCV}, October 2019.

\bibitem{Xian}
Ke~Xian, Chunhua Shen, Zhiguo Cao, Hao Lu, Yang Xiao, Ruibo Li, and Zhenbo Luo.
\newblock Monocular relative depth perception with web stereo data supervision.
\newblock In {\em IEEE CVPR}, pages 311--320, 2018.

\bibitem{ContinousCRFCNN2017}
Dan Xu, Elisa Ricci, Wanli Ouyang, Xiaogang Wang, and Nicu Sebe.
\newblock Multi-scale continuous crfs as sequential deep networks for monocular
  depth estimation.
\newblock In {\em 2017 IEEE Conference on Computer Vision and Pattern
  Recognition (CVPR)}, pages 161--169, 2017.

\bibitem{SAGCNF2018}
Dan Xu, Wei Wang, Hao Tang, Hong Liu, Nicu Sebe, and Elisa Ricci.
\newblock Structured attention guided convolutional neural fields for monocular
  depth estimation.
\newblock In {\em 2018 IEEE/CVF Conference on Computer Vision and Pattern
  Recognition}, pages 3917--3925, 2018.

\bibitem{CADepth2021}
Jiaxing Yan, Hong Zhao, Penghui Bu, and YuSheng Jin.
\newblock Channel-wise attention-based network for self-supervised monocular
  depth estimation.
\newblock In {\em 2021 International Conference on 3D Vision (3DV)}, pages
  464--473, 2021.

\bibitem{Yan2020DHDRINonLocal}
Qingsen Yan, Lei Zhang, Yu~Liu, Yu~Zhu, Jinqiu Sun, Qinfeng Shi, and Yanning
  Zhang.
\newblock Deep hdr imaging via a non-local network.
\newblock {\em IEEE Transactions on Image Processing}, 29:4308--4322, 2020.

\bibitem{HSMNet2019}
Gengshan Yang, Joshua Manela, Michael Happold, and Deva Ramanan.
\newblock Hierarchical deep stereo matching on high-resolution images.
\newblock In {\em IEEE CVPR}, pages 5510--5519, 2019.

\end{thebibliography}

\vspace{-0.3in}
\begin{IEEEbiography}[{\includegraphics[width=1in,height=1.25in,clip,keepaspectratio]{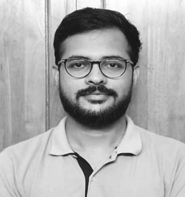}}]{Rohit Choudhary} completed his B.Tech. in Electronics and Communication Engineering, in 2019, from National Institute of Technology Kurukshetra, India. He is currently pursuing M.S. at the Department of Electrical Engineering, Indian Institute of Technology Madras, India. His research interests include 3D Computer Vision, Computational Photography, Deep Learning and 3D Display Technologies. 
\end{IEEEbiography}
\vspace{-0.4in}
\begin{IEEEbiography}[{\includegraphics[width=1in,height=1.25in,clip,keepaspectratio]{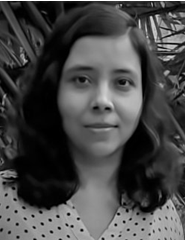}}]{Mansi Sharma} received Ph.D. in Electrical Engineering, in 2017, from IIT Delhi. She received M.Sc. Degree in Mathematics, in 2008, and M.Tech. Degree in Computer Applications, in 2010, from Department of Mathematics, IIT Delhi. She is a recipient of the INSA/DST INSPIRE Faculty award, 2017. Since May 2018, she has been working as an INSPIRE Faculty in the Dept. of Electrical Engineering, IIT Madras. Her research interests include 
Computational Photography, Computational Imaging, Machine Learning, Artificial Intelligence, Virtual and Augmented Reality, 3D Display Technology, Wearable Computing, Visuo-haptic Mixed Reality, Applied Mathematics and Data Science. 
\end{IEEEbiography}

\vspace{-0.3in}

\begin{IEEEbiography}[{\includegraphics[width=1in,height=1.25in,clip,keepaspectratio]{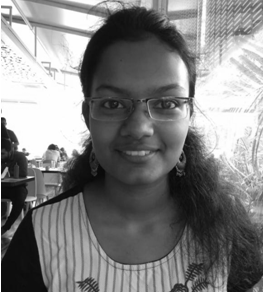}}]{Uma T. V.} is a final year student at Indian Institute of Technology Madras, currently pursuing her Interdisciplinary dual degree – B.Tech. in Mechanical Engineering and M.Tech. in Data Science. Her research interests include Machine Learning, Computer Vision, Autonomous Driving and Clinical Decision Support.
\end{IEEEbiography}

\vspace{-0.4in}

\begin{IEEEbiography}[{\includegraphics[width=1in,height=1.25in,clip,keepaspectratio]{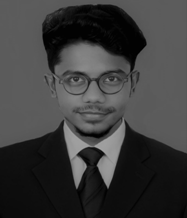}}]{Rithvik Anil} completed his B.Tech. in Mechanical Engineering and M.Tech. in Robotics, in 2021, from India Institute of Technology Madras, India. His research interests include 3D Computer Vision, Autonomous Vehicles and Deep Learning.
\end{IEEEbiography}

\end{document}